\documentclass[lettersize,journal]{IEEEtran}

\usepackage{amsmath,amsfonts}
\usepackage{algorithmic}
\usepackage{algorithm}
\usepackage{array}
\usepackage[caption=false,font=normalsize,labelfont=sf,textfont=sf]{subfig}
\usepackage{textcomp}
\usepackage{stfloats}
\usepackage{url}
\usepackage{verbatim}
\usepackage{graphicx}
\usepackage{cite}
\usepackage{enumitem}
\usepackage[pdfstartview=XYZ,
bookmarks=true,
colorlinks=true,
linkcolor=blue,
urlcolor=blue,
citecolor=blue,
pdftex,
bookmarks=true,
linktocpage=true, 
hyperindex=true
]{hyperref}
\usepackage{orcidlink}

\usepackage[absolute]{textpos}
\begin{document}

\begin{textblock}{181}(17.5,266)
\noindent \scriptsize
\hrule
\vspace{0.3em}
\noindent
\copyright\ 2023 The Authors. This work is licensed under a Creative Commons Attribution 4.0 License. For more information, see https://creativecommons.org/licenses/by/4.0/.
This article has been accepted for publication in IEEE Transactions on Geoscience and Remote Sensing. This is the author's version which has not been fully edited and content may change prior to final publication. Cite as: J. Alatalo, T. Sipola and M. Rantonen, "Improved Difference Images for Change Detection Classifiers in SAR Imagery Using Deep Learning," in \emph{IEEE Transactions on Geoscience and Remote Sensing}, vol. 61, pp. 1-14, 2023, Art no. 5218714, doi: 10.1109/TGRS.2023.3324994.
\end{textblock}

\title{Improved Difference Images for Change Detection Classifiers in SAR Imagery Using Deep Learning}

\author{Janne Alatalo$^{\orcidlink{0000-0001-5515-4419}}$\,, Tuomo Sipola$^{\orcidlink{0000-0002-2354-0400}}$\,, and Mika Rantonen$^{\orcidlink{0000-0002-5320-0853}}$
\thanks{J. Alatalo, T. Sipola, and M. Rantonen are with Jamk University of Applied Sciences, Institute of Information Technology, Piippukatu 2, 40100 Jyv\"asky\"a, Finland (e-mail: \texttt{janne.alatalo@jamk.fi}, \texttt{tuomo.sipola@jamk.fi}, \texttt{mika.rantonen@jamk.fi})}
}

\maketitle

\begin{abstract}
  Satellite-based Synthetic Aperture Radar (SAR) images can be used as a source of remote sensed imagery regardless of cloud cover and day-night cycle. However, the speckle noise and varying image acquisition conditions pose a challenge for change detection classifiers. This paper proposes a new method of improving SAR image processing to produce higher quality difference images for the classification algorithms. The method is built on a neural network-based mapping transformation function that produces artificial SAR images from a location in the requested acquisition conditions. The inputs for the model are: previous SAR images from the location, imaging angle information from the SAR images, digital elevation model, and weather conditions. The method was tested with data from a location in North-East Finland by using Sentinel-1 SAR images from European Space Agency, weather data from Finnish Meteorological Institute, and a digital elevation model from National Land Survey of Finland. In order to verify the method, changes to the SAR images were simulated, and the performance of the proposed method was measured using experimentation where it gave substantial improvements to performance when compared to a more conventional method of creating difference images.
\end{abstract}


\section{Introduction}

\IEEEPARstart{R}{emote} sensing change detection can be used for many purposes, such as damage assessment after a natural disaster~\cite{sublime2019, dong2013, long2014}, detection of forest damages after a storm~\cite{tomppo2021, ruetschi2019}, and monitoring deforestation and glacier melting~\cite{debem2020, sood2021}, to name only a few. Change detection works by comparing two images that have been captured at different dates in the same geographical location and finding the areas that have changed during the time between the acquisitions~\cite{lu2004}. Different platforms can be used to image the terrain, such as airplanes and satellites, however only satellites provide the advantage of continuously monitoring the whole planet~\cite{fu2020}. The revisit time of some satellite systems can be as short as only a few days, and the images are available from anywhere in the planet. This makes the satellite images a useful source of remote sensing data for change detection applications. Some space agencies, such as European Space Agency (ESA), provide some of the satellite images for anybody to download and use~\cite{castriotta2022}. The ease of acquiring the data further facilitates the development of change detection systems that are based on the satellite remote sensing techniques. The images from the satellites are captured using either optical or radar sensors, with radar having the advantage of piercing the cloud layer, thus enabling it to work in various weather conditions~\cite{fu2020}. However, the radar satellites have their disadvantages as well. The resolution of the images is not as good as what the optical instruments can produce. The resolution of the radar images is defined by the antenna length and the frequency band of the radar signal. To enable higher resolution images, the satellites use the synthetic aperture radar (SAR) technique, where the satellite movement over the ground is utilized to synthesize virtual aperture that is longer than the physical antenna on the satellite~\cite{moreira2013}. However, even with the SAR technique the radar images are lower resolution when compared to the optical images. ESA has the Sentinel-1 mission with two SAR satellites that operate on the C-band and have the spatial resolution of around \(5 \times 20\) meters~\cite{sentinelInterferometricWideSwath}. Likewise, speckle noise reduces the quality of the SAR imagery. SAR images always have a grainy look from the speckle, which is random noise that is always present in the images. Despite the shortcomings of the SAR imagery, they are commonly used in remote sensing change detection~\cite{gong2014, zhu2013, zhang2022, li2019}.

\begin{figure*}[t]
  \centering
\includegraphics[width=\textwidth]{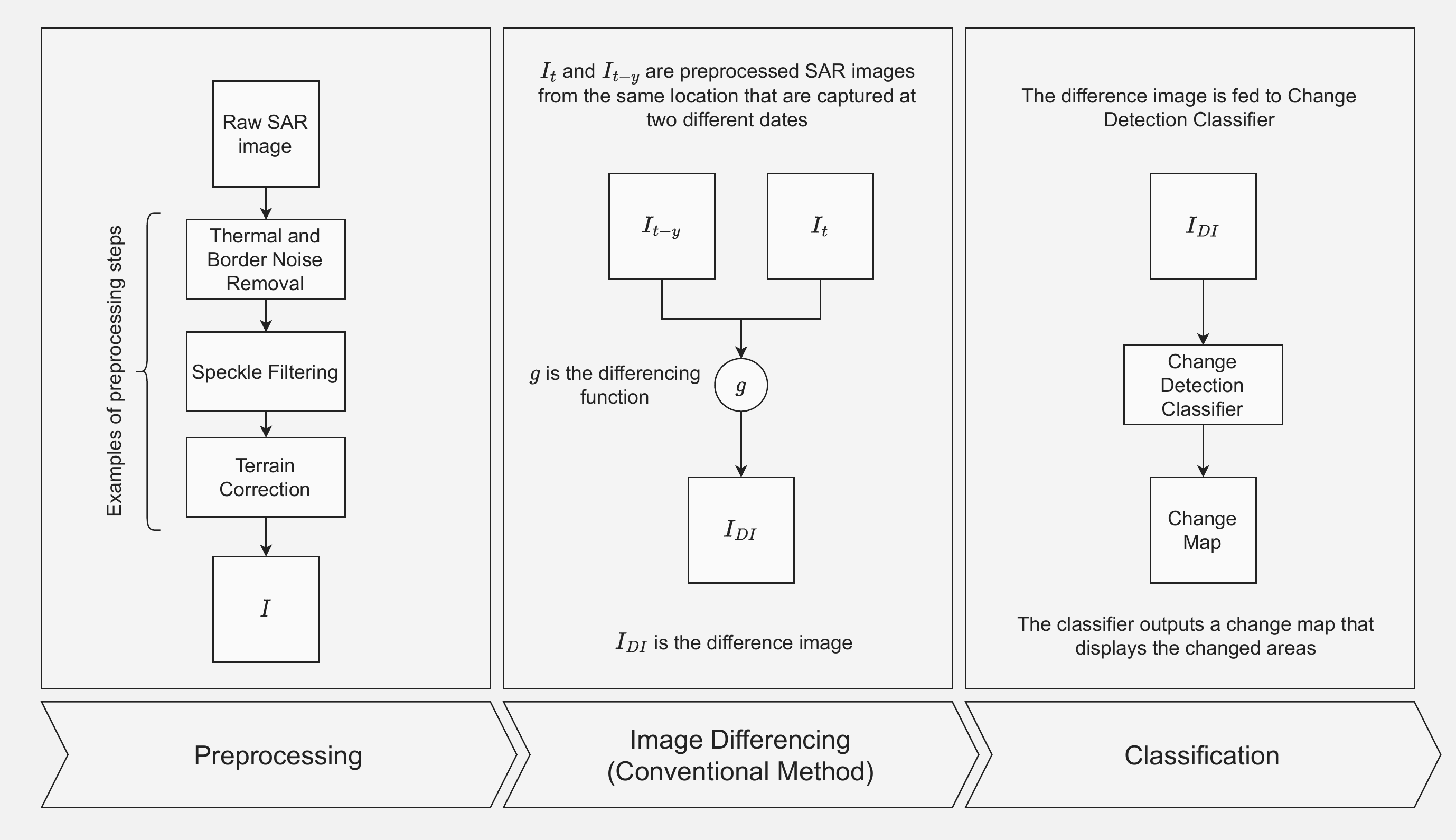}
  \caption{Change detection is often implemented in three distinct steps. The first step is to make the images more comparable to each other using a preprocessing pipeline. The preprocessed images are then used to create difference images (\(I_{DI}\)) using a function \(g\) that is often a algebraic operation, such as subtraction, ratio, or log ratio. The \(I_{DI}\) is then used as an input to a change detection classifier that produces the change map that displays the changed areas. The figure illustrates the conventional method of producing the difference images by using two SAR images that are captured from the location in two different dates.\label{conventional-method}}
\end{figure*}

One approach to implement a change detection system, that is generally used in unsupervised change detection, is to proceed in steps~\cite{bruzzone2002}. Figure~\ref{conventional-method} illustrates this method. The images are first preprocessed to make them comparable among each other. Then, two images from the same location, that are captured at different times, are used to produce a difference image (DI) using an algebraic operation like subtraction, ratio, or log ratio. Finally, the DI is analysed by a classifier algorithm to produce a change map that indicates the changed regions. The preprocessing step is crucial for this method to work well. The issue with the speckle noise is commonly recognized problem with change detection on SAR imagery~\cite{gong2014, zhang2022, li2019}, and to mitigate the issue, noise suppression algorithms are used in the image preprocessing step. However, it is impossible to remove the noise completely, thus the DI also includes noise that causes misclassifications in the classification step. Likewise, other image properties that influence the image comparability have an effect to the quality of the DI. This includes properties such as the satellite orbit direction, incidence angle, and ground moisture content. The satellite does not capture the image from the same angle during every revisit. In case of the ESA Sentinel-1 satellites, the satellite can be flying from North to South, or from South to North, during the image acquisition, and the satellite orbit can be higher or lower with respect to the horizon from the ground perspective between the overflies. The satellite imaging angle influences how the radar signal backscatters from the ground features~\cite{gauthier1998}, which results in that images taken from different imaging angles likely produce lower quality DI than images taken from the same imaging angle. Likewise, ground weather conditions can influence the DI quality. Soil moisture content changes the dielectric constant of the soil, thus changing the backscatter intensity of the radar signal~\cite{srivastava2009}. Images that are taken in similar weather conditions are likely to produce better quality DI when compared to images that are taken in different weather conditions. One solution to improve the DI quality is to favour images with similar acquisition conditions when selecting the images that are used to produce the DI. However, this is not always possible.
Predicting vegetation properties from atmospheric conditions has been identified as one of the potential tasks to benefit from neural networks with spatio-temporal context. 
However, prescriptive assumptions could limit this use. 
The combination of process-based modeling and data-driven machine learning approaches could help when the physical models need support from data~\cite{reichstein2019}.
Adding domain knowledge to the physical layers is a step towards hybrid modeling. Indeed, deep neural networks have been used to extract spatial and frequency features from SAR images. Using these features, the classification of objects or areas in SAR images is also suitable as a deep learning task~\cite{datcu2023}.

The contribution of this paper is a new method to produce better quality difference images. This is achieved by using a neural network-based mapping transformation function preprocessing step that factors in the image acquisition conditions of the SAR images, which improves the comparability of the SAR images. Existing research about SAR image preprocessing has focused on removing speckle noise from the images~\cite{touzi2002, argenti2013}, or correcting the incidence angle variation~\cite{chen2022, kaplan2021}. However, to the best of knowledge of the authors, this is the first time when the comparability of the SAR images is improved by taking in to account the overall image acquisition conditions using a neural network-based preprocessing step. Project code is available on GitHub~\footnote{\url{https://github.com/janne-alatalo/sar-change-detection}}.

\section{Materials and Methods}

\subsection{Proposed Method}

\begin{figure*}
\includegraphics[width=\textwidth]{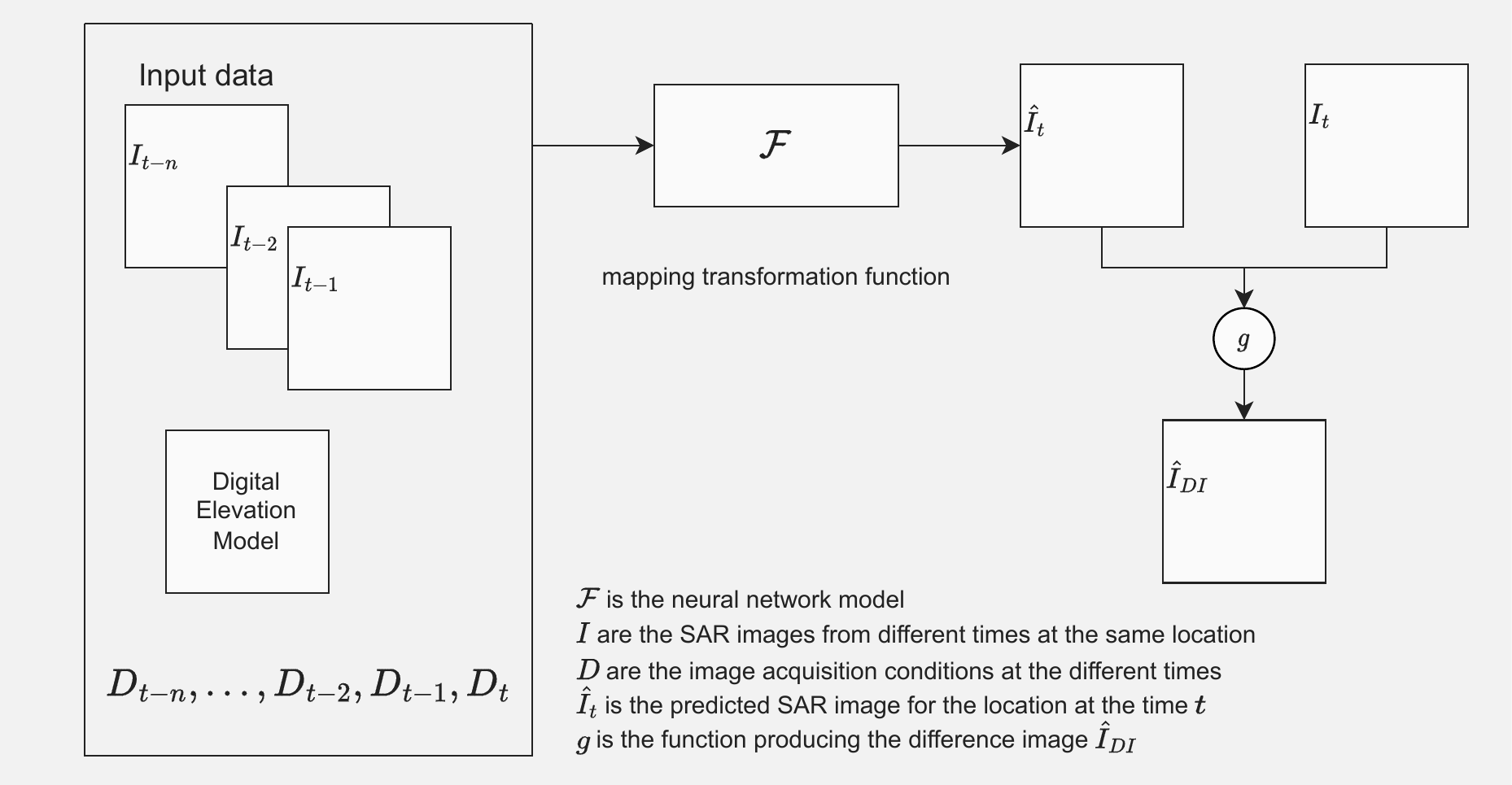}
  \caption{Architectural overview of the proposed method. The neural network-based mapping transformation function fuses the information from previous image acquisitions and predicts what the scene should look like at the imaging conditions of \(I_t\). The model output image \(\hat{I}_t\) and the actual image \(I_t\) is used to produce the difference image \(\hat{I}_{DI}\).\label{mapping-transformation-function}}
\end{figure*}

Figure~\ref{mapping-transformation-function} illustrates the overall architecture of the proposed method. It replaces the conventional method that is illustrated in Figure~\ref{conventional-method} image differencing step. The idea of the proposed method is to improve the SAR image comparability by considering the acquisition conditions of the SAR images. The proposed method utilizes a mapping transformation function that creates artificial SAR images in the requested acquisition conditions. The mapping transformation function \(\mathcal{F}\) is a neural network model that is trained to predict the SAR image \(I_t\) at the time \(t\). The neural network output \(\hat{I}_t\) is the artificial SAR image that is created in the acquisition conditions of \(I_t\), therefore it should be more comparable to the \(I_t\) than previous SAR images from the location that might have been captured in different acquisition conditions.

The model input consists of three distinct features, which are: The previous SAR images from the location; the acquisition conditions of the SAR images (including at time \(t\)); and the digital elevation map from the location. The objective of the neural network model is to learn to replicate the SAR image at the time \(t\). The only information from the time \(t\) in the model input are the image acquisition conditions of the \(I_t\). This means that for the model to be able to replicate the \(I_t\), it needs to learn to map the information contained in the previous SAR images and the digital elevation map to the image acquisition conditions of the \(I_t\). With an ideal model that could perfectly replicate the \(I_t\), the \(\hat{I}_t\) and \(I_t\) would be identical if nothing has changed between the image acquisition of the \(I_{t - 1}\) and \(I_t\), however the \(\hat{I}_t\) would be missing the change if something had changed after the previous image acquisition since the information of the change is not included in the model input data. In practice the SAR images include random noise that is impossible to replicate accurately, and the acquisition conditions are not accurate enough for perfect replication of the \(I_t\), therefore the \(\hat{I}_t\) only approximates the \(I_t\).

The objective function of the model training to produce artificial SAR images is defined as follows:

\begin{equation*}
  \mathcal{L} = \sum_{i = 0}^{N}(h(I_t)_i - h(\hat{I}_t)_i)^2,
\end{equation*}

\noindent
where $h(I)$ flattens the image $I$ by concatenating the pixel rows of the image to one dimensional array. $N$ is the length of the array. The objective function is mean squared error (MSE) between the most recent image and the predicted image.

The intuitive description of the \(\hat{I}_t\) is that the neural network-based mapping transformation function produces a prediction how the \(I_t\) should look like based on previous information about the location and the actual imaging conditions of the \(I_t\). The produced image \(\hat{I}_t\) can be used with the actual image \(I_t\) to create the difference image \(\hat{I}_{DI}\) by using a simple algebraic operation like subtraction, ratio, or log ratio. Generating the difference image is the standard method of conducting change detection, especially when using unsupervised methods~\cite{bruzzone2002}.

Conventional methods of producing the difference image often use only one of the previously captured images with the most recent image to generate the image e.g. \(I_{DI} = g(I_t, I_{t - y})\)~\cite{zhuang2019}. This method has the previously discussed drawbacks of noise and imaging conditions affecting the final difference image quality. By using the proposed mapping transformation function, the predicted image \(\hat{I}_t\) is used in the place of the previously captured image to generate the difference image e.g. \(\hat{I}_{DI} = g(I_t, \hat{I}_t)\). Recall that \(\hat{I}_t\) is a representation of \(I_t\) based on geospatial information from the time \(t - 1\) and earlier, therefore it is missing all the changes that have happened after that time. The predicted image \(\hat{I}_t\) does not contain noise and the mapping transformation function can correct the acquisition condition mismatch between the images, therefore the proposed method should produce better quality difference images when comparing it to the conventional method.

SAR imaging is sensitive to the soil moisture content of the imaged area~\cite{srivastava2009}. Change in the soil moisture level changes the dielectric constant of the soil, and that way changes the SAR backscatter intensity. Often the soil moisture content changes should be ignored by the change detection system. Otherwise, the system would notify changes after every rainy day. This is one of the advantages of the proposed method. By adding weather to the model input acquisition condition parameters, the mapping transformation function can learn to construct the \(\hat{I}_t\) in the actual weather conditions of \(I_t\) and should correctly model the changes in the soil moisture changing the backscatter intensity. Therefore, the false positive changes, that are potentially caused by weather condition changes, should be reduced.

In addition of weather, the acquisition condition parameters also include the imaging angle and identify the satellite that captured the image. A location is imaged by one of the sentinel satellites with an interval ranging from a few days to about a week. The satellite does not capture the image from the same angle every time. The satellite can be in ascending or descending orbit during the image acquisition and the incidence angle can vary between the overpasses. The ascending or descending orbit changes the look direction of the satellite, and that way has a considerable affect to the resulting image. The Sentinel-1 satellites are right-looking. When the satellite is descending from North to South it is imaging to the direction of West, and for ascending passes it is imaging to the direction of East~\cite{mora2016}. Various 3D features, like forest edges, lake banks and hills are sensitive to the look direction, therefore the imaging angle is an important parameter when computing the difference image. When using an image differencing method where only one previous image is used for difference image computation, the imaging angle of the most recent image can restrict what previous images can be used to produce the difference image. Seasonal changes, like foliage growth or change in snow cover, means that the most optimal image for the differencing would be the most recent previous image, however different imaging angles can limit the usage of the most recent images. This problem is not present with the proposed method. The model input includes \(n\) previous images and their imaging angle information. The model output image \(\hat{I}_t\) is produced using the actual acquisition conditions of \(I_t\). The model can use all the information from all \(n\) input images, despite the input including images from different look directions, and the produced image \(\hat{I}_t\) represents an image that is acquired from the same angle as \(I_t\).

\subsection{Neural Network Architecture}

\begin{figure*}[t]
\centering
\includegraphics[width=\textwidth]{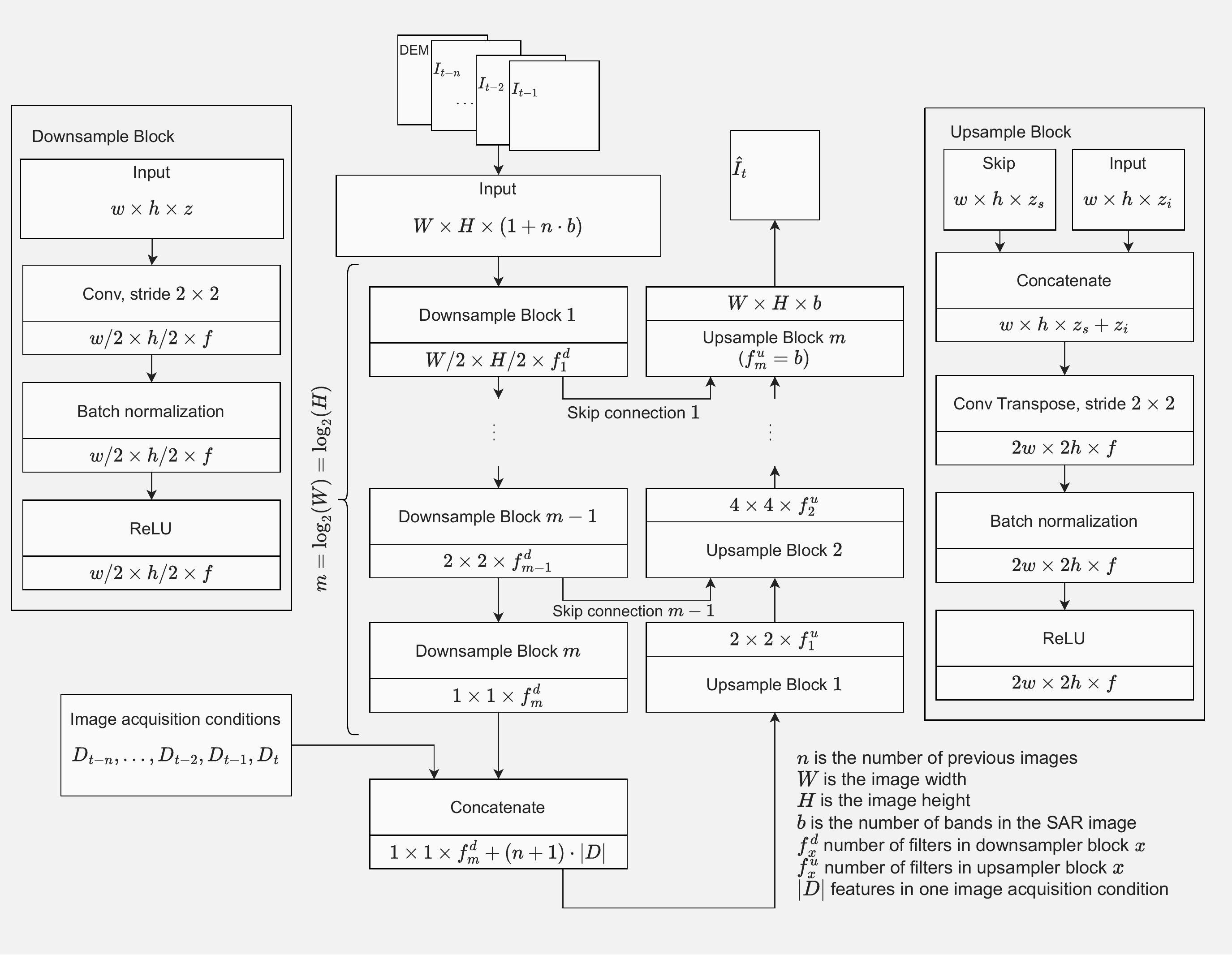}
\caption{The neural network architecture for the mapping transformation function. The architecture is based on the well-known U-Net neural network architecture. The image acquisition conditions are injected to the latent vector between the encoder and decoder.\label{u-net-architecture}}
\end{figure*}

Figure~\ref{u-net-architecture} illustrates the architecture of the neural network-based mapping transformation function. The architecture is based on the well-known U-Net neural network architecture~\cite{ronneberger2015}. The previous \(n\) SAR images, and the digital elevation map (DEM) are stacked to construct the input. The previous images and the DEM are all from the same location. The images are projected to the same resolution and the pixels across the different images are aligned to match the same geographical position. The U-Net architecture is constructed from encoder and decoder units. The encoder takes the input and compresses the input image stack to the latent space by using a set of downsampler blocks that half the input resolution using convolution layers with stride \(2 \times 2\). The encoder stacks enough downsampler blocks so that the input image stack is compressed to \(1 \times 1\) resolution in image height and width dimensions. The image acquisition conditions vector, that contains the information of the acquisitions conditions for the \(n\) input images and the target image, is concatenated to the latent vector as described at the end of section \ref{sec:dataset}. The resulting vector is then fed to the decoder that decodes the vector back to the dimensions of a normal SAR image outputting the \(\hat{I}_t\). The decoder is constructed from upsample blocks that double the width and height dimensions using transposed convolution layers with stride \(2 \times 2\). The decoder has same amount of upsampler blocks as the encoder has downsampler blocks. The number of filters, that are used in the upsampler and downsampler blocks, can be configured for every block individually, except for the final upsample block that has the same number of filters as the SAR image has bands. The encoder and decoder layers are connected with skip connections that help the model in producing the output by not forcing the model to pack all the information to the latent vector. Instead, the information can flow from the input to the output by skipping most of the layers in the architecture. This is a standard method in U-Net style architectures.

\subsection{Dataset}\label{sec:dataset}

A dataset is needed for the training of the neural network-based mapping transformation function. As discussed previously, the mapping transformation function input is composed from the previously taken SAR images; the acquisition conditions of the previous and the most recent SAR image; and the digital elevation map from the location. The objective of the model is to learn to predict the most recent SAR image based on the input, therefore the most recent SAR image is the target in the training dataset. This means that the training dataset does not require any labelled data making the learning process of the proposed method unsupervised and economical to implement. The dataset can be generated directly from available data sources without the need of human labelling for the data. The dataset is available at the Fairdata.fi service~\cite{dataset2023}. 

\begin{figure*}[t]
\includegraphics[width=\textwidth]{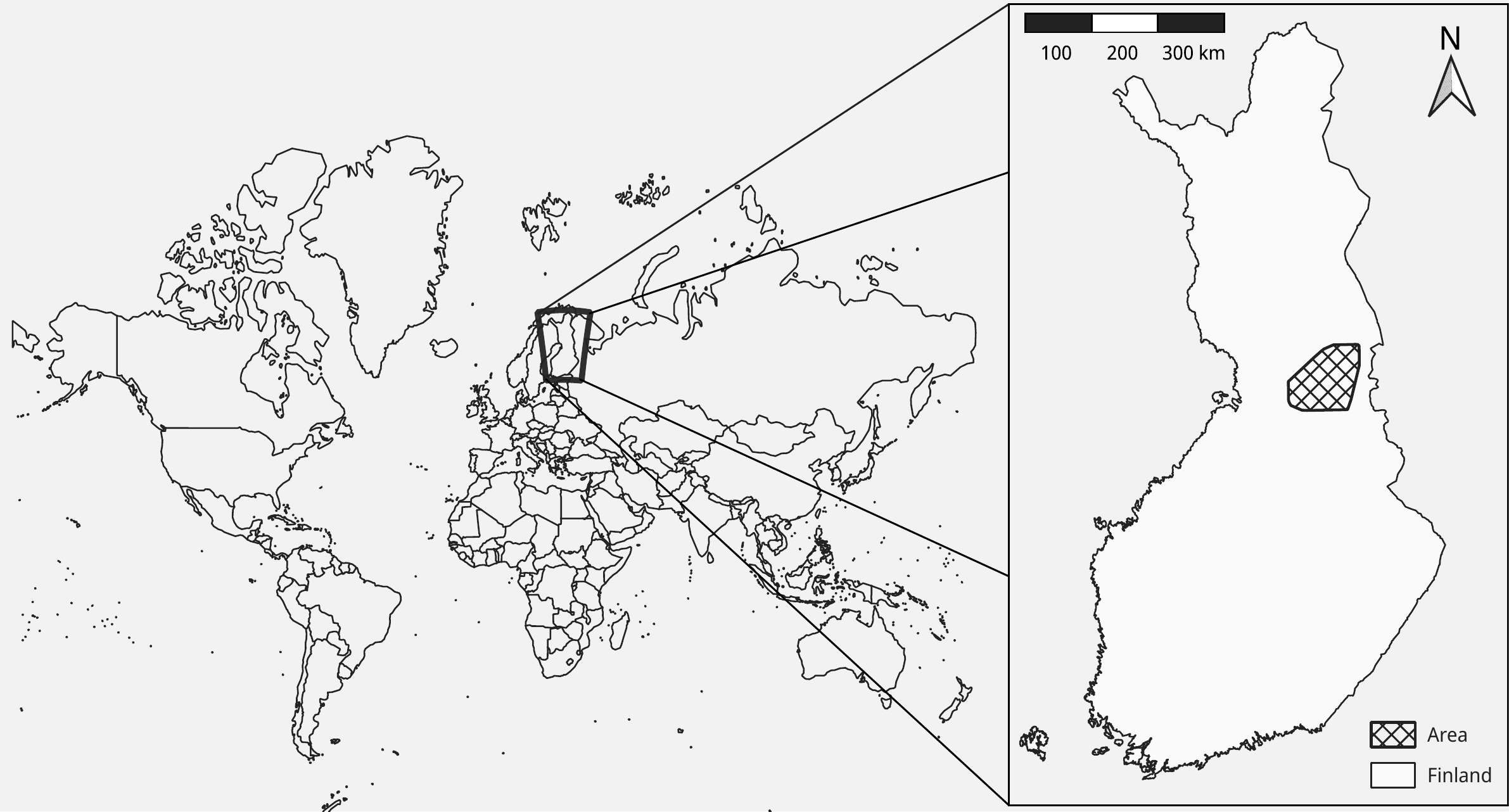}
\caption{The dataset was generated from images acquired from the marked area. The figure contains data from the National Land Survey of Finland Topographic Database~\cite{NLSadm} and data from  @EuroGeographics distributed by Eurostat~\cite{countries}.\label{imaging-area}}
\end{figure*}

The SAR images for the dataset were acquired from the ESA Copernicus Open Access Hub~\cite{sentinel1Data}. High resolution Interferometric Wide Swath (IW) Ground Range Detected (GRD) products were used in this study~\cite{sentinelGRDProduct}. The images were captured between March 2020 and August 2021 from the area illustrated in the Figure~\ref{imaging-area}. All images from the time frame that included the area were downloaded from the Copernicus Open Access Hub. The images were preprocessed using the Sentinel-1 Toolbox from the Sentinel Application Platform (SNAP)~\cite{snap}, by applying the data preprocessing workflow described by Filipponi in~\cite{filipponi2019}. The optional noise filtering step was applied to the dataset using the Refined Lee filter from the SNAP toolkit. The more accurate AUX\_POEORB precise orbit files were used in the Apply Orbit File step. The AUX\_POEORB files are available 20 days after the image acquisition~\cite{podSpecification2022}, and since the processing was done in spring 2022, the more accurate orbit files were available for all images. The proposed workflow in~\cite{filipponi2019} uses the SRTM Digital Elevation Database in the Range Doppler Terrain correction step, however the database does not cover the area from where the dataset was created, therefore the Copernicus 30m Global DEM was used that does cover the area. The SNAP toolkit can automatically download the required DEM files during preprocessing and the Terrain Correction step supports multiple different DEM sources, including the Copernicus 20m Global DEM, thus the change was trivial to implement. The preprocessed images were saved as GeoTIFF files and uploaded to PostgreSQL~\footnote{\url{https://www.postgresql.org/}} database that was using the PostGIS~\footnote{\url{https://postgis.net/}} extension. Using a relational database as the storage backend simplified the dataset generation process since all the data was available in one place and queryable with SQL.

Although the Copernicus 30m Global DEM was used in the SAR image terrain correction preprocessing step, the product was not used for the mapping transformation function input. Instead, we used more accurate DEM from National Land Survey of Finland (NLS). NLS provides the DEM in multiple different resolutions of which the most accurate 2m grid DEM was used~\cite{NLSdem}. The data is open access and distributed under Attribution 4.0 International (CC BY 4.0) license~\footnote{\url{https://creativecommons.org/licenses/by/4.0/}}. The DEM was downloaded in GeoTIFF format and uploaded to the same PostgreSQL database with the SAR images.

As discussed before, the image acquisition condition data included information about the weather when the images were captured. This data was acquired from Finnish Meteorological Institute (FMI) that provides daily weather observations that are interpolated to \(1 \times 1\)\ km grid~\cite{fmiGriddedObservations}. The interpolation method is described by Aalto et al. in~\cite{aalto2016}. The data is distributed in NetCDF format and uploaded once a month. Daily mean temperature, daily precipitation sum, and snow depth data was downloaded from the time range. The daily observations were extracted from the NetCDF files, converted to daily GeoTIFF rasters, and uploaded to the same PostgreSQL database with the SAR images and DEM.

The final data samples were created by sampling random locations from the area and random dates from the time range. For training dataset, the time range was limited to the time before 20th of June in 2021, and for the test dataset the time was limited after the date. An assumption was made that the samples do not have any changes between the acquisitions \(I_{t - 1}\) and \(I_t\). This assumption is not likely true for all of the samples, however the total dataset size is created to be large enough so that the samples that have changes between the two last acquisition should be marginally small fraction of the total dataset and neural networks can adapt to noisy data~\cite{song2022}. The image size was set to \(512 \times 512\) pixels, and number of previous images was set to \(4\). The spatial resolution of a high resolution IW GRD product is \(20 \times 22\) meters, and the images are distributed with \(10 \times 10\) meter pixel spacing~\cite{sentinel1ProductDefinition}. The geographical dimensions of the images were set to \(3 \times 3\)\ km making the pixel size \(3000 / 512 \approx 5.9\) meters. This is higher resolution than the original 10 meter pixel size, therefore the information loss is minimized during processing. For each random location and date, the target SAR image \(I_t\) was the next SAR image from the location that was available after the date. The input SAR images \(I_{t - 4}, I_{t - 3}, I_{t - 2}, I_{t - 1}\) were the SAR images from the four previous acquisitions from the location that were captured before the \(I_t\). The SAR images and the DEM was queried from the PostgreSQL database and the rasters were projected to the same projection window with the same \(512 \times 512\) resolution and \(3 \times 3\)\ km spatial dimensions using GDAL library~\cite{gdalLib}. The \texttt{gdal.Translate} function was used for the projection with nearest neighbor resampling algorithm. After the projection, all pixels were geographically aligned across all images and the images could be stacked to construct the input image stack. The Sentinel-1 satellites use Interferometric Wide swath mode with dual polarization over the land areas thus one SAR image has two bands~\cite{sentinelInterferometricWideSwath}. Both bands are used in all input images and the target image. That makes the input image stack to have \(1 + 4 \cdot 2 = 9\) channels (DEM has one channel and every SAR image has two bands/channels), and the model output image has two bands.

The acquisition conditions were composed from the following features:

\begin{enumerate}
\item	Mean temperature of the acquisition date
\item	Snow depth in the acquisition date
\item	Satellite orbit direction during the acquisition (Ascending/Descending)
\item	Incidence angle
\item	Satellite id (Sentinel-1A or Sentinel-1B)
\item	Precipitation amount in the acquisition date and three previous dates
\end{enumerate}

In addition of imaging conditions, such as weather and imaging angle, the satellite that captured the image is also added to the acquisition condition vector. The satellite id is encoded to \(1\) if the image is captured by the S1A satellite and \(0\) if the image is captured by the S1B satellite. The imaging instrumentation is not necessarily identical in both satellites and the model might learn to use this information to create more accurate images. All other features were scalar values from the acquisition date except for precipitation that is a vector with values for four different days. Since the moisture content of the soil has known effect to the signal, and moisture can linger long times in the soil, it was decided to include the precipitation amounts from multiple days to the acquisition conditions. Taking the precipitation amounts from the previous \(4\) days was a somewhat arbitrary decision with a reasoning that the neural network can learn to ignore the precipitation amounts from previous days if they have no use. The features were flattened to the final vector with dimensionality of \(|D| = 9\).

The final generated dataset had had around \(230,000\) training samples, and around \(9,000\) test samples.

\subsection{Experiment Setup}

The performance of the proposed method was measured using experimentation. The main contribution of this paper is to offer a new strategy for computing the difference image. Existing methods generally use a strategy where the difference image is computed using \(I_{DI} = g(I_{t - y}, I_t)\), where the \(g\) is the differencing function, \(I_{t - y}\) is one of the previous images from the location captured at some previous date, and \(I_t\) is the most recent image from the location. The proposed method uses the neural network output \(\hat{I}_t\) in place of the \(I_{t - y}\) to compute the difference image \(\hat{I}_{DI} = g(\hat{I}_t, I_t)\). The mapping transformation function factors in the imaging conditions of \(I_t\) when generating the \(\hat{I}_t\), therefore the \(\hat{I}_{DI}\) should be higher quality when compared to \(I_{DI}\). The difference image is generally further used in the change detection system to detect the changes by applying a classifier to the difference image. The classifier outputs a change map indicating the pixels that contain the detected changes. By using identical classifier to classify the difference images generated by the two different methods and comparing the classifying accuracy of the resulting change maps, the quality of the two difference images can be measured.

\subsubsection{Change Simulation}

The experiment needs a dataset with known changes so that the accuracy of the change detection classifier can be determined. This is a challenge since only a small number of datasets exists for remote sensing change detection even for optical satellite images~\cite{shi2020}. For SAR images there are only few datasets such as the ones used in the following publications~\cite{luppino2019, wang2022}, however they consist of only few SAR image pairs with a hand labelled change map. Currently there are no large enough SAR datasets for deep learning applications available online~\cite{du2023}.

To avoid the problem with the lack of change detection datasets for SAR images, the decision was made to use simulation to add changes to real SAR images. This technique was used by Inglada and Mercier in~\cite{inglada2007} where they measured the performance of their statistical similarity measure change detection algorithm using simulated changes. The authors used three different methods for change simulation. The techniques were: \textit{offset change}, where the original value was shifted by a value; \textit{Gaussian change}, where the original value was changed by adding zero mean Gaussian noise to the value; and \textit{deterministic change}, where a value was copied from some other location in the image. Likewise, Cui et al. used change simulation for SAR images when they introduced an evaluation benchmark for SAR change detection algorithms~\cite{cui2016}. The change simulation methods in the paper try to replicate changes that are commonly seen in the real world using techniques that correctly resemble the statistical properties of the real world changes. Based on these papers two change simulation methods were devised for this study.

\begin{enumerate}
  \item{\textit{Offset change}: A value is added to the original pixel value. The simulation does not try to replicate any real world change, however it is trivial to implement, and the offset value can be changed to test different offsets.}
  \item{\textit{First-order statistical change}: The statistical distribution of the change area is converted to the statistical distribution of some other nearby geographical feature. This replicates the real world changes more accurately.}
\end{enumerate}

\begin{figure*}
  \centering
    \subfloat[Original image]{
      \includegraphics[width=2.5in]{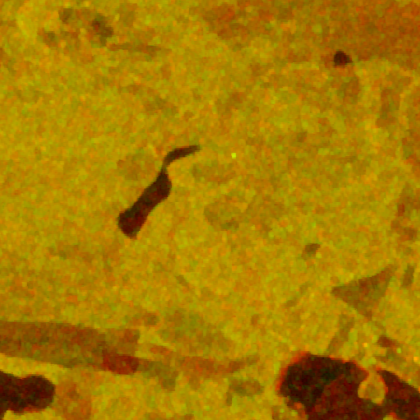}
      \label{orig-img}
    }
    \subfloat[Change mask]{
      \includegraphics[width=2.5in]{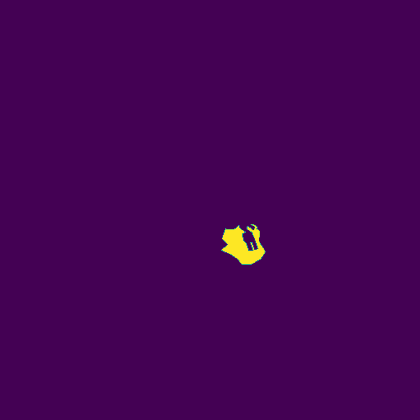}
      \label{change-mask}
    }\\
    \subfloat[Offset change]{
      \includegraphics[width=2.5in]{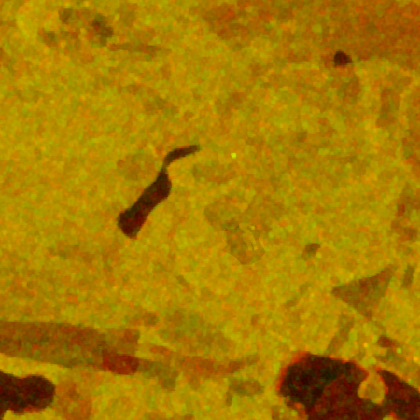}
      \label{offset-change}
    }
    \subfloat[First-order statistical change]{
      \includegraphics[width=2.5in]{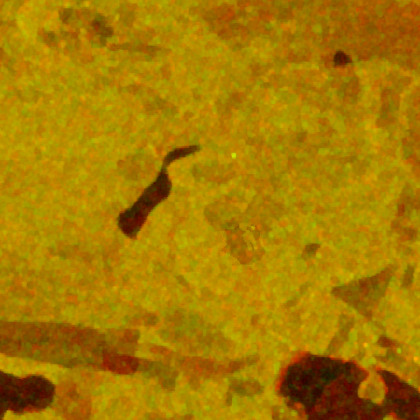}
      \label{first-order-statistical-change}
    }
    \caption{Example of the two simulated change methods. The original image is shown in the Figure~\ref{orig-img}. The SAR images are visualized as RGB image by using red and green channels for the two bands. The blue channel is set to zero. The offset change is \(-2.5\) dB in the image~\ref{offset-change}, that is close to the mean change introduced by the first-order statistical change method in the image~\ref{first-order-statistical-change}.}
  \label{example-simulated-changes}
\end{figure*}

Figure~\ref{example-simulated-changes} illustrates the simulated change methods when applied to an example SAR image. The changes were added to the SAR images by creating a random shape mask and positioning the mask to a random location in the SAR image. The pixel values inside the mask were changed using the selected method. The location of the mask was restricted to forested geographical areas in the SAR image. If the mask location was at forest edge, the mask part that landed outside of the forested area was not changed. The information about different geographical features was acquired from the NLS Topographic Database~\cite{NLSTopoDB}. The database was also utilized in first-order statistical change implementation where the forest area pixel values were changed to follow the statistical distribution of some other geographical feature. The nearest areas of the desired geographical feature type were queried from the database, and the statistical distribution of the pixel values was estimated using a univariate kernel density estimator (KDE) from the statsmodels Python library~\cite{seabold2010statsmodels}. A second univariate KDE model was fitted to the pixel values of all forested area pixels in the SAR image. The mapping of the pixel values was implemented using the method of modifying first-order statistical distribution described in~\cite{cui2016}. The change area pixel values were first mapped to uniform distribution in the interval \([0, 1]\) by using the cumulative distribution function (cdf) of the forest area KDE. After that, the inverse cdf of the second KDE model is applied to the uniformly distributed values, thus mapping them to the distribution of the desired geographical feature.

The simulated change dataset is only needed for validating that the difference images that are generated using the proposed method are higher quality when compared to the difference images that are generated using the conventional method. The simulated changes are not used for training the neural network. Therefore, the simulated changes are added only to the neural network test dataset samples. A random number of changes, ranging from \(0\) to \(3\), were added to each of the samples.

\subsubsection{Difference Image Generation and Change Classifiers}

The quality of the difference images was measured using two different classifiers. The first method is a simple threshold method. A thresholding value is chosen, and the pixels are classified to changed or unchanged based on if the value is smaller or greater than the threshold. This requires that the pixels have scalar values. The scalar valued difference images were produced using the following equations:

\begin{equation}
  \hat{I}_{DI}(x, y) = \sqrt{\sum_b{(I_t(x, y, b) - \hat{I}_t(x, y, b))^2}}
\end{equation}
\begin{equation}
  I_{DI}(x, y) = \sqrt{\sum_b{(I_t(x, y, b) - I_{t - y}(x, y, b))^2}}
\end{equation}

In the equations, \(\hat{I}_{DI}\) is the difference image that is computed using the proposed method, \(I_{DI}\) is the difference image that is computed using the conventional method, \(b\) is the band, and the \(x\) and \(y\) define the pixel location. The different bands are considered as vector dimensions. Pythagorean theorem is used to compute the vector length that is used as the value for the difference image pixel. The threshold method was used as an example of an unsupervised classifier algorithm~\cite{du2023}. The performance of the threshold classifiers was measured using the well known area under curve (AUC) metric that is computed from the receiver operating characteristic (ROC) curve. The metrics were computed to the test partition of the neural network mapping function dataset. The \(\hat{I}_{DI}\) and \(I_{DI}\) difference images were computed for every sample in the test dataset, and the pixels from all samples were used to generate the two datasets that were used to compute the ROC curves and AUC metrics.

The second classifier was the linear support vector classifier (SVC). The support vector classifier was used as an example of supervised machine learning algorithm. The support vector models work with multidimensional data, therefore the difference images were produced using simple subtraction:

\begin{equation}
  \hat{I}_{DI}(x, y, b) = I_t(x, y, b) - \hat{I}_t(x, y, b)
\end{equation}
\begin{equation}
  I_{DI}(x, y, b) = I_t(x, y, b) - I_{t - y}(x, y, b)
\end{equation}

The test dataset from the mapping transformation function training was used to train the classifiers. For each sample, the two difference images were computed, and the pixels from all difference image samples were used to create the two datasets. The first dataset was generated using the pixels from the \(\hat{I}_{DI}\) samples, and the second dataset was generated using the pixels from the \(I_{DI}\) samples. The two datasets were further divided to train and test datasets with a rule that all pixels originating from one image sample end up in the same side of the split. The train test split was also identical for both datasets. The datasets were used to train two instances of the classifier and measure their accuracy.

\section{Results}

\subsection{Training the Neural Network-Based Mapping Transformation Function}

Different neural network parameters were experimented with, and the best results were achieved with the parameters shown in the Table~\ref{model-parameters}. Mean squared error was used as the loss function, and AdamW~\cite{loshchilov2017} was used as the optimizer. The neural network architecture was implemented using TensorFlow deep learning framework~\cite{tensorflow2015-whitepaper}. The training was conducted on one NVIDIA V100 GPU with batch size of \(200\), and training time of around 30 hours.

\begin{table}[!t]
  \caption{Parameters for the neural network architecture}
  \centering
  \begin{tabular}{| p{0.25\columnwidth} | p{0.25\columnwidth} | p{0.25\columnwidth} |}
  \textbf{Block number}	& \textbf{Downsampler (filter size, kernel size)}	& \textbf{Upsampler (filter size, kernel size)}\\
  \hline
  1		& 64, 4             & 512, 4\\
  2		& 128, 4			& 512, 4\\
  3		& 256, 4			& 512, 4\\
  4		& 512, 4			& 512, 4\\
  5		& 512, 4			& 512, 4\\
  6		& 512, 4			& 512, 4\\
  7		& 512, 4			& 256, 4\\
  8		& 512, 4			& 128, 4\\
  9		& 512, 2			& 2, 4\\
\end{tabular}
  \label{model-parameters}
\end{table}
\unskip

\begin{figure*}
  \centering
    \subfloat[Target image \(I_t\)]{
      \includegraphics[width=0.32\textwidth]{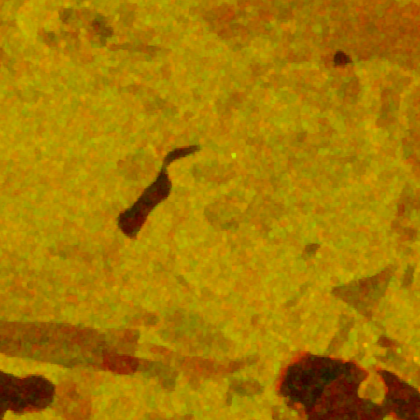}
      \label{latent_change_test_original}
    }
    \subfloat[Difference between \(I_t\) and \(\hat{I}_t\)]{
      \includegraphics[width=0.32\textwidth]{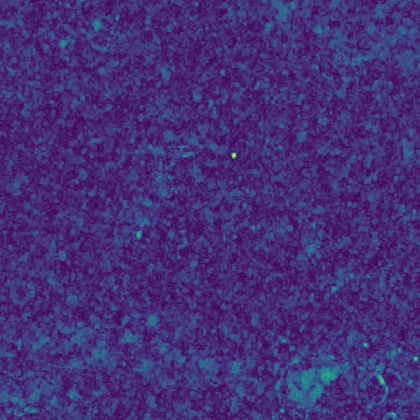}
      \label{latent_change_test_orig_pred_diff}
    }
    \subfloat[Predicted image \(\hat{I}_t\)]{
      \includegraphics[width=0.32\textwidth]{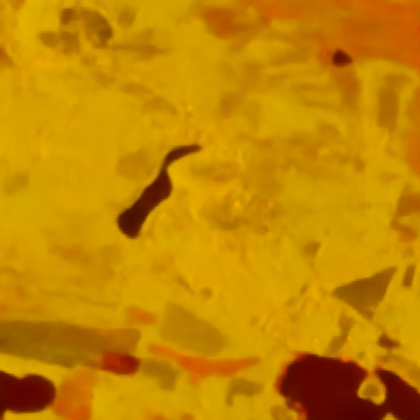}
      \label{latent_change_test_predicted}
    }
    \\
    \subfloat[Ascending orbit direction]{
      \includegraphics[width=0.32\textwidth]{figs/latent_change_test_predicted.png}
      \label{latent_change_test_predicted2}
    }
    \subfloat[Difference between orbit directions]{
      \includegraphics[width=0.32\textwidth]{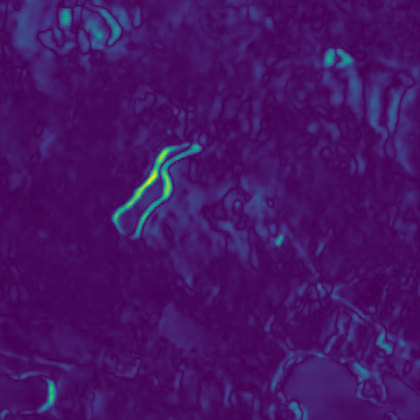}
      \label{latent_change_test_opposite_orbit_diff}
    }
    \subfloat[Descending orbit direction]{
      \includegraphics[width=0.32\textwidth]{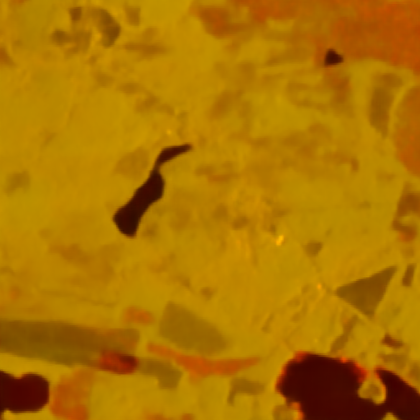}
      \label{latent_change_test_opposite_orbit}
    }
    \\
    \subfloat[Zero rain]{
      \includegraphics[width=0.32\textwidth]{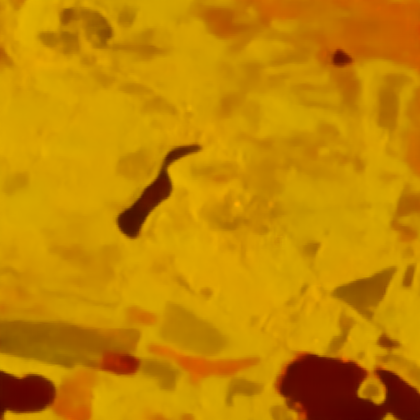}
      \label{latent_change_test_rain_none}
    }
    \subfloat[Difference between the rain conditions]{
      \includegraphics[width=0.32\textwidth]{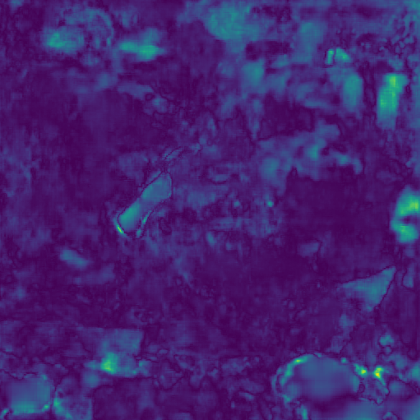}
      \label{latent_change_test_rain_diff}
    }
    \subfloat[Substantial amount of rain]{
      \includegraphics[width=0.32\textwidth]{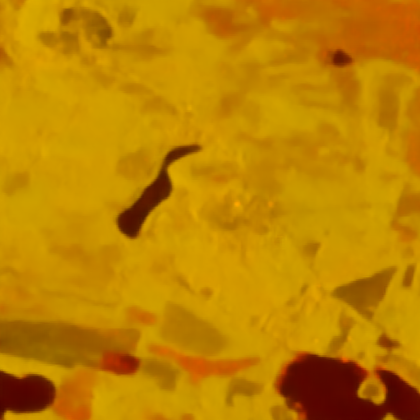}
      \label{latent_change_test_rain_lot}
    }
    \caption{Mapping transformation function outputs with different imaging conditions. The image~\ref{latent_change_test_original} is the original SAR image that is captured from coordinates \(64.919\) lat, \(28.124\) lon in 7th of July 2021. The image~\ref{latent_change_test_predicted} shows the model output \(\hat{I}_t\) when it is trying to predict the \(I_t\). Image~\ref{latent_change_test_orig_pred_diff} shows the difference between the true image \(I_t\) and the predicted image \(\hat{I}_t\). The Images~\ref{latent_change_test_predicted2}, ~\ref{latent_change_test_opposite_orbit}, ~\ref{latent_change_test_rain_none} and ~\ref{latent_change_test_rain_lot} are generated by manually modifying the imaging condition vector \(D_t\). Image ~\ref{latent_change_test_predicted2} has ascending and~\ref{latent_change_test_opposite_orbit_diff} has descending orbit direction. Image~\ref{latent_change_test_opposite_orbit_diff} shows the difference between the different orbit direction images. Identical experiment was conducted by varying the precipitation amount in images \ref{latent_change_test_rain_none} and \ref{latent_change_test_rain_lot}. Image \ref{latent_change_test_rain_diff} shows the difference between the images with the different precipitation amounts.}
  \label{latent-space-change-tests}
\end{figure*}

Figure~\ref{latent_change_test_predicted} demonstrates the model performance for one of the test samples. Figure~\ref{latent_change_test_original} shows the real SAR image that the model tries to predict. Figure~\ref{latent_change_test_orig_pred_diff} illustrates the difference between the real SAR image and the model output with a heat map where lighter color indicates a greater error. The predicted image is very close to the real SAR image except for lack of noise that is purely random and impossible for the model to predict. Likewise, the lower right corner of the image has an area that has greater error in the prediction. The error is located in a lake, therefore the error can be a result of waves that are likewise impossible to predict.

The proposed method depends on that the mapping transformation function adapts the predicted image \(\hat{I}_t\) based on the imaging conditions of \(I_t\). To verify that the model genuinely uses the image acquisition conditions to produce the \(\hat{I}_t\), the model was experimented to produce outputs with manually modified imaging condition vector \(D_t\). Figure~\ref{latent_change_test_predicted2} and Figure~\ref{latent_change_test_opposite_orbit} image pair illustrates model outputs where the \(D_t\) is modified to have opposite orbit directions. Figure~\ref{latent_change_test_opposite_orbit_diff} illustrates the difference between the images. The lake banks and the upper left corner of the image, where there is a small hill, have large differences between the two generated images. All locations, where there are greater differences between the images, are 3D features. The Sentinel-1 satellites have different look directions on ascending and descending orbit directions. Therefore, the scattering of the radar signal is different and the difference is most noticeable on 3D features. Since the differences are so clearly located on the 3D features in the image the model is clearly factored in the orbit direction when generating the output. This verifies that the imaging conditions are used by the model to produce the \(\hat{I}_t\) in the imaging conditions of \(I_t\).

The same experiment was conducted by modifying the precipitation amounts in Figure ~\ref{latent_change_test_rain_none} and Figure~\ref{latent_change_test_rain_lot}. The difference between the generated images is shown in the Figure~\ref{latent_change_test_rain_diff}. This time the difference between the generated images is focused on swamp, meadow, and agricultural land areas in the image. The forest areas have only small differences between the images. In forest areas, the radar signal is scattered back by the forest foliage where the moisture does not affect the scattering properties as much as the open areas. In open areas, the radar signal hits the ground where the soil moisture content is altered more by the rain, thus changing the backscatter intensity. This experiment suggests that the model uses the precipitation information correctly when generating the output image.

\subsection{Identifying the Best Conventional DI Strategy}

The conventional method of computing the difference image is to use one of the previous SAR images that is captured at some preceding date with the most recent image to produce the difference image \(I_{DI} = g(I_{t - y}, I_t)\). There are multiple different strategies when selecting the previous image. The simplest strategy is to select the previous image that is preceding the image that was captured most recently. This strategy has the advantage that the least amount of time has elapsed between the images, therefore the number of natural changes, like foliage growth or soil moisture changes, are minimized. However, the problem is that the previous image has very likely different incidence angle and it might have been captured from different orbit direction (ascending/descending). To make sure that we compare the proposed method to the best conventional method, three different previous image selection strategies were compared to identify the best strategy. The threshold classifier was used to compare the quality of the difference images that were produced using the different strategies. The strategies have different trade offs between the elapsed time and imaging angle:

\begin{enumerate}[leftmargin=*, labelindent=1em, label=Method \arabic*:]
  \item{Closest incidence angle and the same orbit direction.}
  \item{Most recent previous image with the same orbit direction.}
  \item{Most recent previous image preceding the target image (\(I_{t - 1}\)).}
\end{enumerate}

\begin{figure}[t]
  \centering
\includegraphics[width=\columnwidth]{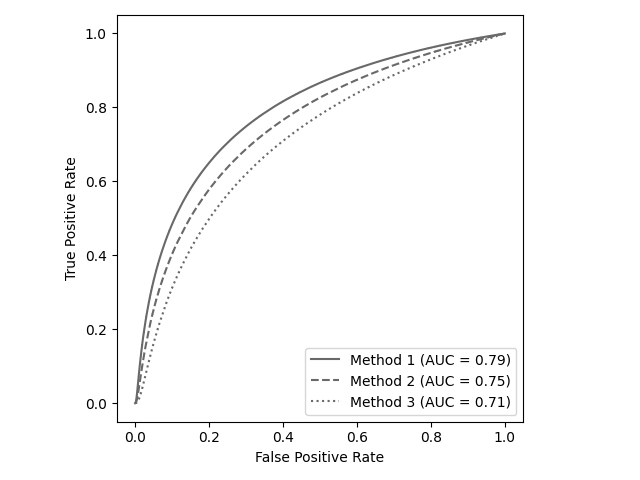}
  \caption{Comparison of different previous image selection strategies when using the traditional method of computing the difference image.\label{selection-methods-roc}}
\end{figure}

Figure~\ref{selection-methods-roc} illustrates the comparison of the three different methods using ROC curve plots. Table~\ref{selection-methods-auc-table} shows the results in a list format by displaying the AUC metrics. The strategy where the previous image is captured from the same orbit direction and has the closest incidence angle with the \(I_t\) is the best \(I_{t - y}\) selection strategy. From this on forward, the Method \(1\) is always used when referring to the conventional method of computing difference image.

\begin{table}[h]
  \caption{Conventional DI Strategy Results}
  \centering
  \begin{tabular}{p{0.4\columnwidth} | p{0.4\columnwidth}}
    \textbf{\(I_{t - y}\) Selection Method}	& \textbf{AUC} \\
  \hline
  Method 1		& 0.79 \\
  Method 2		& 0.75 \\
  Method 3		& 0.71 \\
\end{tabular}
  \label{selection-methods-auc-table}
\end{table}
\unskip

\subsection{Proposed Method vs. Conventional Method}

\subsubsection{Threshold Classifier}

\begin{figure}[H]
  \centering
\includegraphics[width=\columnwidth]{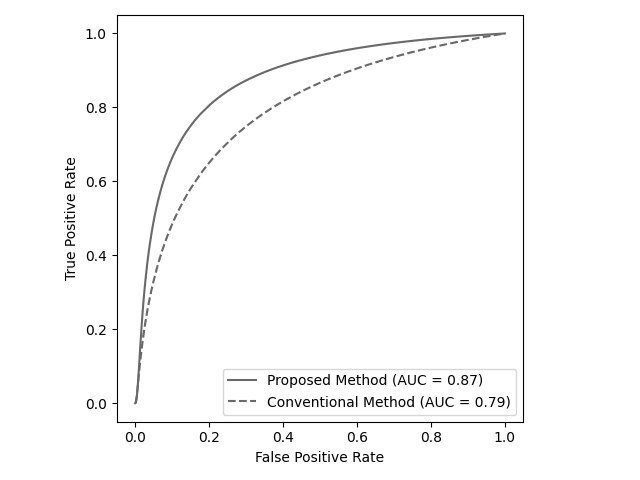}
  \caption{ROC curve for the two threshold classifiers when applied to the dataset with simulated changes using the offset change method.\label{offset-change-threshold-classifier-roc}}
\end{figure}

Figure~\ref{offset-change-threshold-classifier-roc} illustrates the ROC curve plots for the two threshold classifiers when measuring the quality of the difference images generated with the two methods. In this experiment, the changes are simulated to the dataset using the offset change method. The AUC metrics from the experiment is also shown in a list format in the first row of the Table~\ref{experiment-results-table}. The simulated shift is \(-2.5\) dB in the change area, which represents a considerable change. In the real world, this could be a change where the forest is clear cut, making it smoother, and that way reducing the backscatter intensity. The threshold classifier that is using the difference images that are produced using the proposed method is clearly better. This indicates that the proposed method generates better quality difference images.

\begin{table}[h]
  \caption{Experiment results for the threshold models}
  \centering
  \begin{tabular}{p{0.25\columnwidth} | p{0.25\columnwidth} | p{0.25\columnwidth}}
    \textbf{Dataset}	& \textbf{Proposed method AUC} & \textbf{Conventional method AUC} \\
  \hline
    Shift change	& 0.87 & 0.79 \\
    Statistical change & 0.73 & 0.67 \\
\end{tabular}
  \label{experiment-results-table}
\end{table}

\begin{figure}[H]
  \centering
\includegraphics[width=\columnwidth]{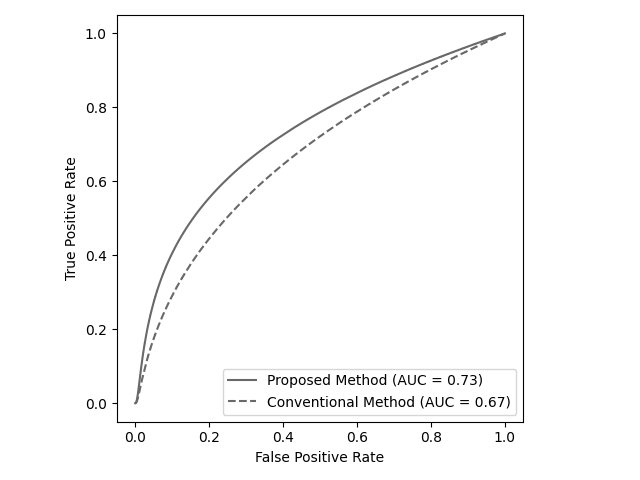}
  \caption{ROC curve for the two threshold classifiers when applied to the dataset with simulated statistical changes.\label{statistical-change-threshold-classifier-roc}}
\end{figure}

Figure~\ref{statistical-change-threshold-classifier-roc} illustrates the results of the same experiment when it is repeated to the simulated change dataset using the statistical change method. The change areas are simulated to emit the backscatter intensity of nearby forest areas that are not as densely wooded making this more realistic representation of real changes in the forest. The mean backscatter intensity change varied from around \(-0.5\) dB to \(-2.5\) dB in the change areas depending on the sample. The AUC metrics from the experiment is also shown in a list format in the second row of the Table~\ref{experiment-results-table}. Both classifiers have considerably worse performance, however the proposed method is still better performing. The overall poor performance is to be expected with the threshold classifiers. It is the simplest possible classifier working in single pixel level without having any kind of visibility to the neighbouring pixels. Furthermore, the changes can be small in the simulated change dataset that is created using the statistical change method.

\subsubsection{Support Vector Classifier}

The experiments were repeated with the SVC model to the same two datasets. The linear kernel SVC implementation \texttt{LinearSVC} from Scikit-learn library~\cite{scikit-learn} was used to conduct the experiment. Linear kernel SVC was chosen due to large dataset size. Other kernel types were tested, however they did not scale to the large number of samples. The samples were normalized using the Scikit-learn \texttt{StandardScaler} to ease the model convergence. Table~\ref{svc-result-table} displays the results from the experiments. The proposed method is clearly superior to the conventional method in both experiments. The performance in the statistical change dataset is considerably worse when compared to the shift change dataset. However, this is to be expected with the similar loss of accuracy in the threshold classifier experiments. This experiment uses supervised learning with labeled dataset which should improve the results when comparing to the threshold classifier. However the SVC is still very simple classifier that performs the classification at pixel level without any visibility to the neighbouring pixels, thus the accuracy scores are mediocre at best. Still, achieving high accuracy score was not the goal of the experiment. Instead, the experiment is comparing the accuracies of the two classifiers and the results from this experiment support the findings from the threshold classifier experiments. The proposed method clearly produces higher quality difference images.

\begin{table}
\caption{Experiment results for the SVC models.}
\centering
  \begin{tabular}{p{0.25\columnwidth} | p{0.25\columnwidth} | p{0.25\columnwidth}}
\textbf{Dataset}	& \textbf{Proposed method accuracy}	& \textbf{Conventional method accuracy}\\
\hline
  Shift change		& \(0.89\)			& \(0.81\)\\
  Statistical change		& \(0.75\)			& \(0.70\)\\
\end{tabular}
  \label{svc-result-table}
\end{table}
\unskip

\subsubsection{Model Without the Weather Data}

\begin{figure}[t]
  \centering
  \includegraphics[width=\columnwidth]{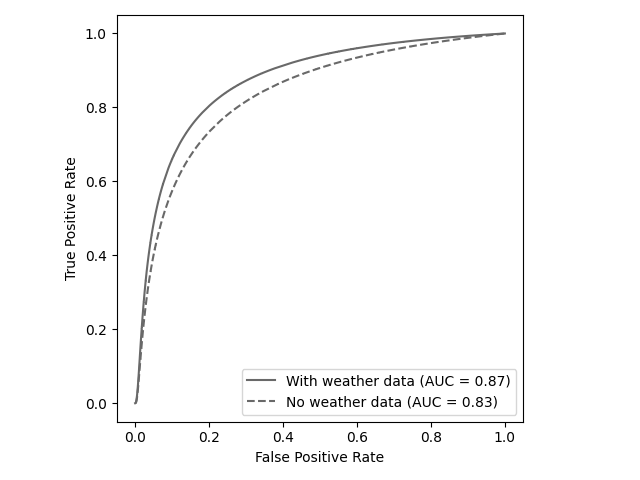}
  \caption{Threshold classifier ROC when used with mapping transformation function that is trained without the weather data. The solid line is the same line from the Figure~\ref{offset-change-threshold-classifier-roc} that is added to the plot to help visual comparison of the results.\label{no-weather-threshold-classifier-roc}}
\end{figure}

The dataset creation for this project was a major undertaking which complicates the adaption of the proposed methodology since the model needs to be trained to every location where it is used. Finnish Meteorological Institute provides the interpolated weather data for the features we used in this study that are available in locations inside the borders of Finland. However, equivalent data sources are not necessary available in other countries. Therefore, we experimented how the neural network based mapping transformation function works without the weather data. The model training pipeline was modified to drop the weather data during training and inference, thus the acquisition conditions consisted only from incidence angle, satellite orbit direction, and satellite id. Figure~\ref{no-weather-threshold-classifier-roc} illustrates the results from the experiment. The experiment used simulated changes with \(-2.5\) dB shift and exact same model hyper parameters with the results that are illustrated in Figure~\ref{offset-change-threshold-classifier-roc}, thus the result is directly comparable. The resulting AUC metric is higher at \(0.83\) when comparing to the conventional method at \(0.79\), however the result is worse when comparing to the model that has visibility to the weather data with AUC metric of \(0.87\). We can conclude that the proposed methodology can be used also without weather data, and it achieves measurable improvement over conventional method. However, to achieve the best performance, the model requires the weather data in addition of the other imaging condition features.

\subsubsection{Feature Ablation Study}

An ablation study measured the impact of each feature to the accuracy of the SVC using both the shift change dataset and the statistical change dataset. 
Table~\ref{ablation-result-table} shows the accuracies of the classifier when the neural network \(\mathcal{F}\) is trained with a dataset where one of the features is removed. Satellite orbit direction and precipitation are two of the most important features, because dropping them decreased accuracy. Satellite id and snow depth were the least important.

\begin{table}
\caption{Ablation study impact to SVC model accuracies.}
\centering
\begin{tabular}{p{0.3\columnwidth} | p{0.17\columnwidth} | p{0.20\columnwidth}}
\textbf{Dropped feature}  & \textbf{Shift change dataset} & \textbf{Statistical change dataset}\\
\hline
Mean temperature          & 0.88 & 0.75 \\
Snow depth                & 0.89 & 0.75 \\
Satellite orbit direction & 0.87 & 0.74 \\
Incidence angle           & 0.88 & 0.75 \\
Satellite id              & 0.89 & 0.76 \\
Precipitation             & 0.87 & 0.74 \\
\end{tabular}
\label{ablation-result-table}
\end{table}

\section{Discussion}

The experiment results show that the proposed method produces higher quality difference images than the conventional method. Since the output from the proposed method is a difference image, many of the existing change classification techniques may benefit from the method without any modifications. The techniques generally use the conventional method for producing the difference image, however it is a completely separate step from classification, and thus could be replaced with the proposed method without changes to the classification step. Some methods do not use the difference image computation step, instead they accept the two images directly to the model to carry out the classification. Even with these techniques the usage of the proposed method could be beneficial. In these cases, the earlier image (\(I_{t - y}\)) is replaced with the \(\hat{I}_t\), thus giving the classification model better understanding about what the scene should look like in the correct image acquisition conditions.

This study did not experiment with the more advanced change detection classifiers since the simple classifiers were enough to prove that the proposed method is better than the conventional method. However, the clear improvement in classification accuracy with the simple methods could indicate that similar improvement can be achieved with the more advanced methods. 

The use of simulated changes to measure the performance of the method was a necessary compromise caused by the lack of existing change detection datasets suitable for training the neural network. The simulated changes are not realistic enough to draw conclusions about how much the proposed method would improve the change detection performance in real world. However, the experiments with the simulated changes indicate a substantial performance improvement potential.

The downside of the proposed method is that the mapping transformation function is a neural network model that requires a training dataset and considerable amount of processing power for training. The dataset creation is a complex operation that combines data from multiple data sources. Some of the sources that were used in this study are available only for geographical locations inside Finland, such as the interpolated weather data from Finnish Meteorological Institute. The model requires training data from the locations it is used at inference time which complicates the adaption of the method outside of Finland. However, many of the data sources very likely have equivalents available in other geographical locations, therefore the adoption is not impossible. Even a global training dataset could potentially be constructed, which could make the training of a universal model possible. The recent advances in neural network architectures with natural language processing and image generation have shown that the models can learn from impressive amounts of data. The model training is unsupervised, meaning it does not require labelled data, thus the creation of such a dataset could be possible. Our experiment with a model that did not see the weather data in the input shows that the method achieves measurable improvement over the conventional method even when the model has information only about the imaging angle and the satellite. That data is available in the SAR images when they are downloaded from the ESA open access portal, thus simplifying the dataset creation considerably. However, without the weather data the mapping transformation function cannot generate accurate enough SAR images to achieve the same accuracy metrics as the model with the weather information. 
The ablation study suggests that model training could be simplified since some of the features are found to be less important, and therefore can be dropped from the training data.

\section*{Funding}

This research was conducted at the Institute of Information Technology of Jamk University of Applied Sciences as part of \textit{Data for Utilisation -- Leveraging digitalisation through modern artificial intelligence solutions and cybersecurity} project funded by the Regional Council of Central Finland/Council of Tampere Region and European Regional Development Fund (grant A76982), and \textit{coADDVA - ADDing VAlue by Computing in Manufacturing} project funded by REACT-EU Instrument as part of the European Union’s response to the COVID-19 pandemic (grant A77973), and \textit{Finnish Future Farm} project, co-funded by the European Union and the Regional Council of Central Finland with the Just Transition Fund (grant J10075).

\section*{Data Availability}
The Sentinel-1 SAR imagery is available to download free of charge from Copernicus Open Access Hub~\cite{sentinel1Data}. The weather data is available free of charge from Finnish Meteorological Institute~\cite{fmiGriddedObservations}. The digital elevation map and topographic database are available free of charge from the National Land Survey of Finland open data service~\cite{NLSdem, NLSTopoDB}.  Links to the download sites are listed in the references. The derived dataset 
can be downloaded from the Fairdata.fi service~\cite{dataset2023}. The computer code to produce the results is available at \url{https://github.com/janne-alatalo/sar-change-detection}.

\section*{Acknowledgments}
This work uses modified Copernicus Sentinel data 2020-2021. The authors thank Mr. Eppu Heilimo for feedback on the draft manuscript. The authors wish to acknowledge CSC - IT Center for Science, Finland, for hosting the dataset.

\bibliographystyle{IEEEtran}
\bibliography{refs.bib}

%
%

\end{document}